\documentclass{article}

\usepackage[preprint,nonatbib]{neurips_2024}
\usepackage{graphicx}
\usepackage[utf8]{inputenc} % allow utf-8 input
\usepackage[T1]{fontenc}    % use 8-bit T1 fonts
\usepackage{hyperref}       % hyperlinks
\usepackage{url}            % simple URL typesetting
\usepackage{booktabs}       % professional-quality tables
\usepackage{amsfonts}       % blackboard math symbols
\usepackage{microtype}      % microtypography
\usepackage{xcolor}         % colors
\usepackage{tabularx}
\usepackage{subfig}
\usepackage{amsmath}
\usepackage{authblk}

\usepackage{array}
\newcolumntype{L}{>{\raggedright\arraybackslash}X}

\title{Digital Typhoon Dataset V3: Benchmarking for Multi-spectral Extension and Analyzing Systematic Bias in Intensity Regression}
\author[1,2]{Asanobu Kitamoto}
\author[3,1]{Victor Le Louarn}
\author[4,1]{Noelia Falczuk}
\author[5,1]{Ngoc Anh Tong}
\author[6,1]{Anja Veleva}
\affil[1]{National Institute of Informatics, Japan}
\affil[2]{Typhoon Science and Technology Research Center, Yokohama National University, Japan}
\affil[3]{University of Artois, France}
\affil[4]{University of Buenos Aires, Argentina}
\affil[5]{Hanoi University of Science and Technology, Vietnam}
\affil[6]{RWTH Aachen University, Germany}

\begin{document}
\maketitle
\begin{abstract}
This paper presents the Digital Typhoon Dataset V3, a new version of the longest typhoon satellite image dataset spanning nearly 50 years, designed to benchmark machine learning models on long-term spatio-temporal data. The latest addition to Dataset V3 is multispectral satellite data, enabling multi-channel models. We investigate data-splitting strategies and present benchmarking results for the Northern and Southern Hemispheres. 
Experimental results show that, under the current dataset construction and experimental setting, multi-channel inputs do not clearly outperform single-channel inputs for pressure regression. Among the evaluated model architectures, image types, and spectral channels, ResNet-18, full-resolution images, and Channel 1 generally provide the best single-channel performance for pressure regression. Performance also differs between the WP and AU datasets by regression target, with lower RMSE on WP for pressure regression and on AU for wind regression. These results highlight the importance of considering dataset-specific and task-specific characteristics when interpreting benchmark performance. 
Additionally, we analyzed systematic bias in intensity regression, which leads to underestimation of intensity for strong typhoons. We tested imbalance-aware weighting strategies, including inverse-density weighting inspired by BalancedMSE, and showed that these approaches are not sufficient to remove the bias. 
The dataset is publicly available at \url{http://agora.ex.nii.ac.jp/digital-typhoon/dataset/} and \url{https://github.com/kitamoto-lab/digital-typhoon/}.
\end{abstract}

\section{Introduction}
The Digital Typhoon Dataset was first introduced in November 2023 as a machine learning dataset for tropical cyclones \cite{kitamoto2023digital}. It contained typhoon-centered satellite images with best-track data from the Japan Meteorological Agency. It was designed to support a range of machine learning tasks, including intensity forecasting. The dataset is unique in its long temporal coverage, spanning 1978 to 2025 and encompassing nearly 50 years of typhoon observations. 

We also extended this dataset to a few additional dimensions. Version 2 of the dataset, released in November 2024, extended it to the Southern Hemisphere, enabling comparisons of machine learning models across both hemispheres to understand regional differences \cite{kitamoto2024machine}. The dataset size also increased from 54GB to 77GB by adding data from the latest season and a new hemisphere. 

This paper introduces Version 3 of the dataset. It extends the dataset in the spectral dimension by adding more satellite channels. Versions 1 and 2 offered only an infrared channel; Version 3 added a second infrared channel, a water vapor channel, and a near-infrared channel to capture different aspects of tropical cyclones. As a result, the dataset size increased to 164.9GB for the WP and 66.4GB for the AU, for a total of 231.3GB. 

These channels raise research questions about whether multi-spectral models outperform single-channel models. To test this hypothesis, we trained basic vision models to evaluate single- and multi-channel model performance on this dataset. Additionally, we explored strategies to analyze systematic bias in intensity regression, which leads to underestimation of intensity for
strong typhoons.

\section{Digital Typhoon Dataset V3}
\label{sec-dataset}
\subsection{Temporal Dimension}
\label{sec:temporal}
As in the V2 dataset, we provide separate datasets for the two hemispheres: the Northern Hemisphere dataset (WP: Western Pacific) and the Southern Hemisphere dataset (AU: Around Australia). For both hemispheres, we extended the temporal dimension with annual updates. For the Northern Hemisphere, we added data for the 2025 typhoon season (January-December 2025) using best-track data from the Japan Meteorological Agency (JMA). For the Southern Hemisphere, we also added data for the 2025 cyclone season (July 2024--June 2025) using best-track data from the Bureau of Meteorology (BoM), Australia.

As a result, we extended the dataset period and increased the number of tropical cyclones (TC) and images, as summarized in Table~\ref{tab:two-basins} and Table~\ref{tab:number-two-basins}. 

\begin{table}[t]
    \caption{Characteristics of the WP and AU datasets.}
    \label{tab:two-basins}
    \centering
    \begin{tabularx}{\textwidth}{LLL} 
    \toprule
       Feature        & WP dataset  &  AU dataset \\ 
      \midrule
     Temporal coverage & 1978-2025 & 1979-2025 \\
     Spatial coverage & 100E to 180E in the Northern Hemisphere & 90E to 160E in the Southern Hemisphere  \\
    Season & January to December & July (the previous year) to June \\
     Best track & Japan Meteorological Agency (JMA) & Bureau of Meteorology (BoM), Australia \\
       \bottomrule
    \end{tabularx}
\end{table}

\begin{table}[t]
    \caption{Number of TCs and Images for the WP and AU datasets for V3-2025.R1.}
    \label{tab:number-two-basins}
    \centering
    \begin{tabular}{lrrrr} 
    \toprule
      Ch. & WP TCs & WP images &  AU TCs & AU images \\ 
      \midrule
     1 & 1,169 & 202,986 & 511 & 76,211 \\
     2 & 744 & 142,583 & 339 & 58,865 \\
     3 & 744 & 142,591 & 339 & 58,842 \\
     4 & 502 & 96,611 & 223 & 40,238 \\
            \bottomrule
    \end{tabular}
\end{table}

\subsection{Spectral Dimension}
\label{sec:spectral}

 \begin{table}[t]
    \caption{Channel numbers assigned in the dataset and the duration of each channel across the Himawari satellite series.}
    \label{tab:channels}
    \centering
    \begin{tabularx}{\textwidth}{llLLLLL} 
    \toprule
         Ch. & Type & Himawari-1/2/3/4 & Himawari-5 & GOES-9 & Himawari-6/7 & Himawari-8/9 \\
      \midrule
       1 & Infrared & 10.5-12.5 & 10.5-11.5 (11.0) & 10.2-11.2 (10.7) & 10.3-11.3 (10.8) & $11.2 \pm 0.2$ \\
     2 & Infrared & - & 11.5-12.5 (12.0) & 11.5-12.4 (11.95) & 11.5-12.5 (12.0) & $12.35 \pm 0.3$ \\ 
     3 & Water Vapor & - & - & 6.5-7.0 (6.75) & 6.5-7.0 (6.75) & $6.25 \pm 0.37$ \\
     4 & Near Infrared & - & - & - & 3.5-4.0 (3.75) & $3.85 \pm 0.22$ \\
         \bottomrule
    \end{tabularx}
\end{table}

A new addition to the dataset is the spectral dimension; namely, multiple channels from the Himawari satellite series. 
The first satellite, Himawari-1, has two channels: infrared and visible. In later generations, the number of channels increased by adding more channels with narrower bandwidths. The latest satellites, Himawari 8 and 9, each have 16 channels. 
More channels (wavelengths) are driven by observational and research needs to capture different aspects of the Earth system. More channels may enable improved analysis and forecasting of tropical cyclones. 

Sensor characteristics have evolved across satellite generations, so matching channel specifications is not straightforward. For example, in the early days of the Himawari series, there was only one infrared channel, and its wavelength range was later split into two. In this case, we could define channel identities in several ways. Here, we prioritize dataset continuity and assign the same number to channels that overlap in wavelength. Under this policy, we assign the first infrared channel to Channel 1 and the second infrared channel to Channel 2, which appeared after Himawari-5. As a result, we map the channel number to wavelengths available for each satellite as summarized in 
Table~\ref{tab:channels}. As this table shows, the wavelength definitions for each channel differ across satellites, but the wavelength ranges overlap across satellite generations. Using this channel number, the Digital Typhoon Dataset V1 and V2 contained only Channel 1, whereas V3 includes new channels: 2, 3, and 4. 

Because Channel 1 and Channel 2 are next to each other in the spectrum, 
they exhibit very similar spectral characteristics, and the difference between the two channels, rather than their raw values, is used to classify cloud types. In the following benchmarking tasks, we used raw values as input data. An alternative approach is to use the difference between the two channels, but we leave it for future work. 

Channel 3, called Water Vapor, was introduced after GOES-9. It captures moisture in the upper troposphere and shows moisture-transport dynamics. It exhibits smoother patterns than other channels because it samples only the upper troposphere, owing to atmospheric transparency. 

Channel 4, called Near Infrared, spans the electromagnetic spectrum that overlaps the visible and infrared regions. At night, it is close to infrared, but during the day, it incorporates information from sunlight. This diurnal pattern may confound machine learning models for tasks unrelated to diurnal patterns.

Users may wonder why we did not include visible channels in the dataset, as they have been observed throughout the Himawari series. We decided not to include visible channels for the following reasons. First, because visible channels measure sunlight reflectance, they are meaningful only during the day. The raw value depends on the solar zenith angle. They peak around noon local time but approach zero at night. These strong diurnal patterns complicate model evaluation because the amount of available data differs between day and night. Second, visible channels have historically been more susceptible to noise; over extended periods, they become significantly contaminated and therefore unusable for quantitative analysis. For these reasons, we exclude visible channels from the Digital Typhoon Dataset V3, but we may add them in a future release.  

\begin{figure}[t]
    \centering
    \includegraphics[width=\textwidth]{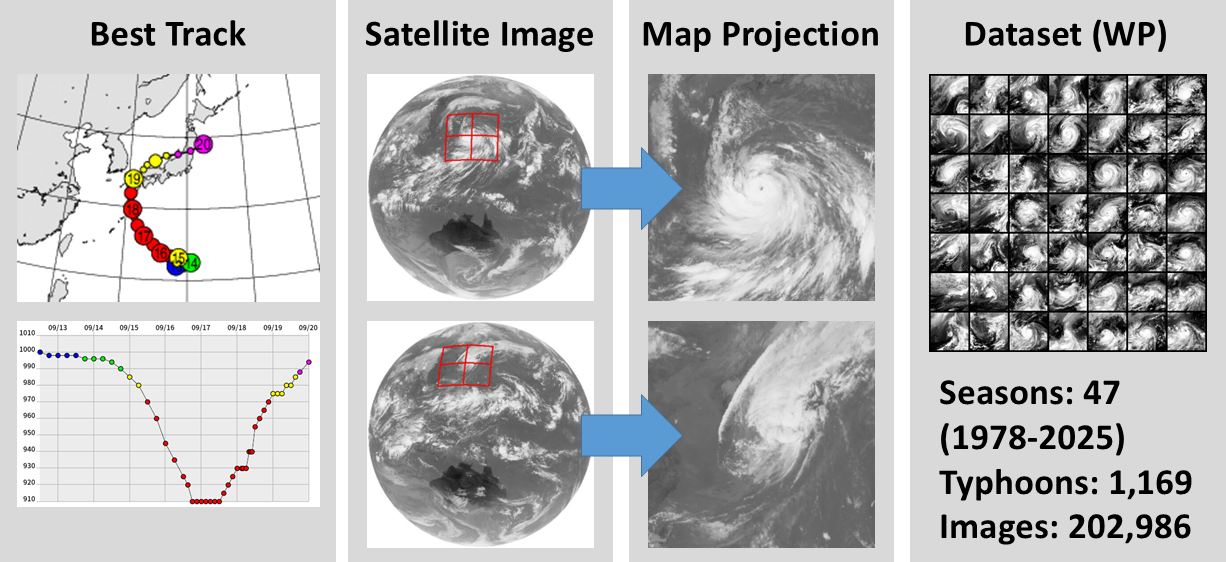}
    \caption{Overview of the data-processing pipeline.}
    \label{fig:overview}
\end{figure}

To create this dataset, we used the same data-processing pipeline used for the V1 and V2 datasets, as shown in 
Figure~\ref{fig:overview}. Please refer to the papers for the V1 \cite{kitamoto2023digital} and V2 \cite{kitamoto2024machine} datasets for details. 

\subsection{Updates on Pyphoon}
The Python-based software library \texttt{Pyphoon} was also updated to \texttt {Pyphoon3}, which includes the following features.

First, it can handle both multi-channel and single-channel datasets. Compared with single-channel datasets, multi-channel datasets must handle cases in which some channels are missing at the same observation time. We require Pyphoon3 to check whether all channels are available at each observation time; if not, we discard that observation time without filling the gap. This results in smaller training and test sets for a multi-channel dataset than for a single-channel dataset.  

Second, we updated the data format. We renamed a variable from 'Temperature' to 'Data' because not all channels measure
temperature. For example, Channel 3 represents water vapor observations, and Channel 4 represents near-infrared observations. We also modified the dataset's directory structure. Previously, we organized the dataset into AU and WP directories, where we stored the images and metadata. However, in dataset V3, we first divided the dataset into channel directories 1 to 4, then into hemisphere directories AU and WP. We chose this organization to support single-channel and multi-channel tasks with the same directory structure, and we adapted Pyphoon3 to accommodate these changes. 

\texttt{Pyphoon3} is downloadable from 
\url{https://github.com/kitamoto-lab/pyphoon3/} and related information is available at 
\url{https://github.com/kitamoto-lab/digital-typhoon/}. 
All experiments in this paper can be reproduced with Pyphoon3 3.0.0 and V3-2025.R1 (available upon request).

\subsection{Dataset Availability}
\label{sec:dataset}

The Digital Typhoon Dataset is available on the official page \url{http://agora.ex.nii.ac.jp/digital-typhoon/dataset/}, or the version-specific page \url{http://agora.ex.nii.ac.jp/digital-typhoon/dataset/V3/},
under an open data license: the Creative Commons Attribution 4.0 International (CC BY 4.0) License. The following is a recommended data citation \cite{dt-dataset}. 

\begin{quote}
    Digital Typhoon Dataset V3 (National Institute of Informatics) 
    doi: \url{https://doi.org/10.20783/DIAS.664}
\end{quote}

The dataset's DOI (Digital Object Identifier) does not change with the version, and we will use the same DOI (10.20783/DIAS.664) for future releases. Although we will phase out datasets V1 and V2 to avoid confusion, we can provide them upon request when reproducibility is required. 

From now on, we will use Semantic Versioning as follows. First, the major version will be used for major updates, such as format changes or adding new data types. Second, we will add an annual update suffix to indicate the last season. 
Third, minor dataset patches are included in the annual update cycle.
As a result, the V3 dataset including the 2025 season is named V3-2025. The next annual update, which will include data for the 2026 season, will be named V3-2026.
Fourth, we add the suffix .R<n> to a version name when we
correct the data within the same annual update. The experiments in
this paper used the first release called V3-2025.R1, and we refer to this dataset as "V3," omitting the suffix for brevity. 

\subsection{Data Corrections in V3-2025.R2}
\label{sec:correction}

While preparing the dataset, we found images in Channels 1--3 of V3-2025.R1 that should not
have been included in the dataset. Part of this problem dates back to Dataset V1 and therefore also
concerns users of Datasets V1 and V2; the rest is in Channels 2 and 3, which are new in V3 and have
not been distributed before. We corrected these images and, following the rule above, named the
corrected release V3-2025.R2. The three problems and their corrections are as follows.

The first two problems produced blank images (a single value). The digital counts of the original
satellite images were always available, but converting them into brightness temperatures in the h5
files requires both a conversion table and recalibration coefficients that correct it, and these could
not be prepared for some observations.

First, for 1,463 Channel 1--3 images from MTS-2 (2009--2012), GMS-2 (1984), and GMS-4 (1989--1990),
the conversion tables were available, but the recalibration coefficients were missing on the
observation date. This is why the Channel 1 images already present in V1 and V2 are missing. In
V3-2025.R2, we used the coefficients of the most recent earlier date with coefficients, or for
observations before the first such date of a satellite, that first date; the difference between the
dates is 1--8 days for 903 images, 163--188 days for 133 GMS-2 images (1984), 216--222 days for 390
MTS-2 images (2009), and 1--10 days for 37 GMS-4 images (1989--1990).

Second, for 2,579 Channel 2 and 3 images from GMS-5 (1995--2003) and GOES-9 (2003--2005), a bug in
the conversion software did not copy the conversion tables of these channels for some data formats,
and for 14 Channel 2 and 3 images from GMS-5 at 02 UTC, the conversion tables were missing in the
source data. In V3-2025.R2, we fixed the bug and converted the 2,579 images with the correct
conversion tables, and for the 14 images we used the conversion table of the observation one hour
earlier (01 UTC on the same day).

Third, the dataset excludes images in which more than 30\% of the area is missing, but a bug in the
detection of missing areas let 46 GOES-9 images on AU (2003--2004; 23 observation times in Channels
1 and 2) pass this check; some of them were already present in Dataset V2. In V3-2025.R2, we fixed
the bug and removed these images.

In total, 4,102 images in V3-2025.R1 were affected: V3-2025.R2 regenerated 4,056 images and removed
46 images. Of these, 579 images were already in V1 (WP) and 616 in V2 (577 on WP and 39 on AU), about
0.3\% of the images. The effect of the substituted coefficients and conversion tables on the accuracy
of the brightness temperatures is not known; we provide the list of these 1,477 images with the
dataset (\texttt{substituted\_conversion.csv}) so that users can exclude them if necessary. Pyphoon3
can also flag images without usable data and exclude them on request (\texttt{exclude\_flagged}).

All benchmarks in this paper were computed on V3-2025.R1. However, to avoid confusion, we distribute
only V3-2025.R2; V3-2025.R1 is available upon request when reproducibility is required. Note that
Tables 2, 4 and 5 give the counts of V3-2025.R1; V3-2025.R2 has 23 fewer images in each of AU-1 and
AU-2. None of the affected images are in the 2024--2025 evaluation seasons, so the evaluation sets
of the SS-L2 split are identical in the two releases; only the training sets differ (at most about
1\% of the training images of a channel). To assess how this affects the benchmarks, we retrained
the baseline model on V3-2025.R2 for five representative configurations (Appendix~\ref{app:r2}).
The median evaluation RMSE improved by about 0.1~hPa on WP, while the differences on AU were within the
variability across runs, so the conclusions of this paper, and likely those of the V1 and V2 papers,
also hold for the corrected data.

\section{Use Cases of the Digital Typhoon Dataset}
\label{sec:usecases}

Since the release of the Digital Typhoon Dataset~\cite{kitamoto2023digital} and its
multi-basin extension~\cite{kitamoto2024machine}, the dataset has been used in
ways that go well beyond the benchmark tasks defined in the original papers. To understand users' needs, we group the published uses into five categories.

\paragraph{Testbed for general-purpose models and toolkits.}
The dataset has been adopted as one task among many in benchmark suites that
evaluate models not designed for tropical cyclones (TCs): video foundation
models (12-hour central-pressure forecasting task)~\cite{hasson2025scivid}, any-sensor Earth-observation foundation
models for wind speed estimation task~\cite{waldmann2025panopticon}, multimodal learning in Earth science~\cite{wang2025omniearth}, uncertainty-quantification
toolboxes~\cite{lehmann2024lightning} and generic video-prediction models~\cite{gmg2026}. It is also shipped as a
built-in dataset in the TorchGeo library~\cite{stewart2025misc}. In this
role, the dataset is valued less for its meteorological content than for being
a long, homogeneous, openly licensed spatio-temporal image sequence with
scalar labels.

\paragraph{Intensity estimation and forecasting.}
The largest group of TC-specific studies targets intensity. Examples include
multi-task estimation~\cite{yan2025idol}; rapid
intensification prediction model using temporal-differential attention~\cite{tdari2024}; phase-based, physics-informed sequence-to-sequence neural network models that train separate networks per lifecycle phase~\cite{phase2025};
physics-informed positional encoding that injects timestamps and coordinates
into vision--language models for multimodal forecasting~\cite{li2025pipe};
quantum-enhanced LSTMs~\cite{quantum2025}; and hybrid CNN--LSTM--random forest
classifiers with synthetic data balancing~\cite{smote2026}, as well as
smaller-scale studies~\cite{gabor2024,ogallo2025multitask}.
A recurring observation across these works is the difficulty of the intense,
data-sparse regime: several papers explicitly attribute underestimation of
extreme intensity to sample imbalance~\cite{phase2025},
and a recent benchmark built on the Digital Typhoon convention reports that
vision foundation models rely on visual shortcuts that collapse for intense
TCs~\cite{yao2026tcbench}. 
This motivates our analysis of systematic bias in intensity regression in Section~\ref{sec:improving-extreme}.

\paragraph{Cloud-image forecasting and generation.}
The hourly image sequences are used
for
physics-enhanced cloud image prediction~\cite{physenh2025}, dual-branch
spatio-temporal prediction~\cite{fusion2026}, and diffusion-based
nowcasting with deterministic--probabilistic ensembles with cross-hemisphere generalization~\cite{tcsatdiff2026}. Conditional diffusion models have used the images
as conditioning input to estimate ERA5 atmospheric variables~\cite{ling2024cddpm}.
The Multi-scale Residual Deformable Attention Model is proposed for satellite-image super-resolution~\cite{mrdam2025}.

\paragraph{Track analysis and forecasting.}
Track studies mostly use the best-track coordinates rather than the images,
for example, ConvLSTM models fused with reanalysis fields~\cite{axioms2023}
and regional studies of typhoon patterns in the Philippines~\cite{stormsignals2025}.

\paragraph{Meteorological, climatological and environmental analysis.}
Beyond machine learning, the best-track records are used as a data source in
climatology and environmental science, such as the northward shift of
landfall over the western North Pacific~\cite{northward2021}, raindrop-size
distributions of South China Sea TCs~\cite{rsd2025}, and the assessment of sea
surface chlorophyll-a in the Seto Inland Sea to typhoon-driven changes~\cite{chl2026}. The dataset is also cataloged alongside best-track
archives and cyclone trackers in disaster-resilience and storm-surge
work~\cite{ogc2024,stormnet2026}, and compared with related datasets built from
other satellites~\cite{lu2025fysatty,huang2025tropicyclonenet,chen2024terra,li2025cllmate}
and in surveys~\cite{chen2023foundation,shi2025survey}.

Two trends stand out. First, most machine-learning uses still rely on the
western North Pacific data alone; the southern-hemisphere data introduced in
V2 has so far been used mainly as a held-out set for cross-hemisphere
generalization~\cite{tcsatdiff2026,gmg2026}. Second, the dataset is
increasingly used to probe general-purpose models rather than to build
TC-specific ones, which makes a well-understood, meteorologically grounded
reference on intensity estimation
all the more useful.

\section{Benchmarking the Digital Typhoon Dataset V3}
\label{sec:benchmarking}

\subsection{Dataset Definitions}
\label{sec:dataset-definitions}

\subsubsection{Single-Channel Datasets}
A single-channel dataset contains all available observations from one satellite channel for a given hemisphere. For example, the Northern Hemisphere dataset containing Channel 1 observations is denoted WP-1, while WP-2, WP-3, and WP-4 denote the corresponding datasets for Channels 2, 3, and 4. The same naming convention applies to the Southern Hemisphere, resulting in AU-1, AU-2, AU-3, and AU-4. Table~\ref{tab:number-two-basins} reports the number of tropical cyclones and images available for each single-channel dataset.

\subsubsection{Multi-Channel Datasets}

We construct a multi-channel dataset from the temporal intersection of the four single-channel datasets. An observation is retained only when Channels 1, 2, 3, and 4 are simultaneously available at the same observation time. Consequently, the multi-channel dataset is smaller than any individual single-channel dataset and has a shorter temporal coverage. The resulting multi-channel dataset contains 502 tropical cyclones and 96,579 observations for WP, and 223 tropical cyclones and 40,236 observations for AU. The intersection is slightly smaller than the Channel 4 dataset because we retain only observations with all four channels available.

In the benchmarking experiments, we considered three input configurations: Channel 1 + 2, Channel 1 + 3, and Channel 1 + 2 + 3. These configurations all use the same underlying multi-channel dataset defined by the four-channel intersection; they differ only in the subset of channels provided as input to the model. We always include Channel 1 because it has the longest temporal coverage and is the reference channel in the Digital Typhoon Dataset V3. We selected Channel 2 and Channel 3 to evaluate the contribution of complementary spectral information. We did not use Channel 4 as a model input because of its distinct characteristics discussed in Section~\ref{sec:spectral} and because the single-channel model for Channel 4 performs poorly, as we show later.  

\subsection{Methodology}
\label{sec:methodology}
Benchmarking evaluates model performance across different spectral channels, image types, and model architectures.

We first introduce the terminology used in this paper. First, we compare base model architectures between ResNet-18 and ResNet-50, two widely used baseline vision models \cite{he2015deepresiduallearningimage}. Although they are not the latest models, we continue using them because these models were used to benchmark datasets V1 and V2. Second, we compare three image types; "Full" denotes the full-resolution image ($512\times 512$) provided in the dataset; "Resized" denotes a resized version of the full-resolution image to ($224\times 224$); and "Cropped" denotes a cropped version centered on the image to ($224\times 224$), preserving the original resolution. This definition matches that in our previous paper \cite{kitamoto2023digital,kitamoto2024machine}. 

We then compare spectral channels to test our hypothesis that multi-channel (multi-spectral) models trained on multi-channel datasets can outperform single-channel models trained on single-channel datasets, because the former can account for cross-channel dependencies during training. 
We first assess single-channel models and compare their performance across channels, image types, and model architectures. This lets us identify the best-performing choices across these experimental dimensions. Second, we compare single- and multi-channel models and evaluate whether incorporating additional spectral information provides a performance advantage under the current dataset construction. We evaluate performance separately for the WP and AU datasets and for pressure and wind regression, as these represent dataset- and task-specific conditions rather than experimental choices.

Based on these settings, we evaluated the regression task primarily on pressure, with wind included for single-channel comparison. We assessed performance separately for WP and AU. We repeated each experiment five times using the same training and evaluation partitions and report the mean RMSE and standard deviation across the five runs.

\subsection{Data Splitting}
\label{sec:data-splitting}
For training and evaluation, we consider two approaches: season-level splitting (SS) and sequence-level splitting (SQ). We also introduce notation to specify the splitting parameters. The first is random selection, which assigns a percentage of the dataset to the evaluation set. For example, retaining $20\%$ of the dataset for evaluation is denoted R20. The second is the last selection, which specifies the number of units in the evaluation set. For example, retaining the last two seasons for evaluation is denoted L2.

Throughout the paper, we primarily use the SS-L2 split for the Digital Typhoon Dataset V3. This means we reserve the last two seasons, 2024 and 2025, as the held-out evaluation set, and use all earlier seasons for training. 
In our experimental protocol, \emph{evaluation set}, \emph{validation set}, and \emph{test set} refer to the same held-out subset of the data. 
We use the held-out evaluation set to monitor model performance during training and report final benchmark results, but we do not use it to fit any model parameters or weighting scheme; we compute these on the training set alone. We therefore use the term \emph{evaluation set} throughout the paper to avoid implying a separate validation set.
We made this choice to ensure consistency across experiments and to simulate operational scenarios where historical data is used to analyze and forecast recent or real-time data. 

Table~\ref{tab:data-split-wp} and Table~\ref{tab:data-split-au} show the dataset size, training set, and evaluation set for the SS-L2 split (2024--2025) for each channel. Note that Channel 1 of the AU dataset contains 169 images that cannot be used in the pressure regression task and are filtered and ignored during data loading by the Pyphoon3 library. The tables reflect this filtering.

In the SS-L2 split, the evaluation set is fixed to the two most recent seasons, so its size is nearly the same across all channels. However, because each channel has different temporal coverage (see Table~\ref{tab:channels}), the percentages of the evaluation set vary across channels, which should be considered when interpreting inter-channel performance comparisons. The SQ-R split reduces the effect of different percentages but is still significantly affected by different dataset sizes across channels.

\begin{table}[t]
\caption{SS-L2 split (2024--2025) for the WP dataset (V3-2025.R1).}
\label{tab:data-split-wp}
    \centering
    \begin{tabular}{lrrr} 
    \toprule
Ch. & Total images & Train images & Evaluation images (\%)\\
\midrule
1 & 202,986 & 192,956 & 10,030 ($4.94\%$) \\
2 & 142,583 & 132,555 & 10,028 ($7.03\%$) \\
3 & 142,591 & 132,554 & 10,037 ($7.04\%$) \\
4 & 96,611 & 86,551 & 10,060 ($10.41\%$)\\
\bottomrule
\end{tabular}
\end{table}

\begin{table}[t]
\caption{SS-L2 split (2024--2025) for the filtered AU dataset (V3-2025.R1).}
\label{tab:data-split-au}
    \centering
    \begin{tabular}{llll} 
    \toprule
Ch. & Total images & Train images & Evaluation images (\%)\\
\midrule
1 & 76,042 & 70,274 & 5,768 ($7.58\%$) \\
2 & 58,865 & 53,097 & 5,768 ($9.80\%$) \\
3 & 58,842 & 53,074 & 5,768 ($9.80\%$) \\
4 & 40,238 & 34,501 & 5,737 ($14.26\%$)\\
\bottomrule
\end{tabular}
\end{table}

We plan to maintain the same data-splitting scheme for future dataset updates. For example, when we add the 2026 season, the evaluation set will shift to 2025 and 2026, with the 2025 season overlapping with the previous evaluation period. In the future, we could also consider an SQ-L1 split in which machine learning models are trained on all historical data while excluding observations from currently active tropical cyclones. However, we could also separate validation and test sets to make better design choices, such as selecting hyperparameters and training duration. Future work includes exploring other data splitting methods for evaluation. 

\subsection{Optimizer selection}
\label{sec:optimizer}
The choice of optimization algorithm significantly affects training dynamics and, consequently, model performance. In our previous benchmarking on the V1 and V2 datasets, we ran all experiments with the Stochastic Gradient Descent (SGD) optimizer. Although SGD is widely adopted, we observed slower convergence and greater sensitivity to the learning rate, which affected training stability and reproducibility across experimental settings.

\begin{figure}[t]
    \centering
    \includegraphics[width=0.95\linewidth]{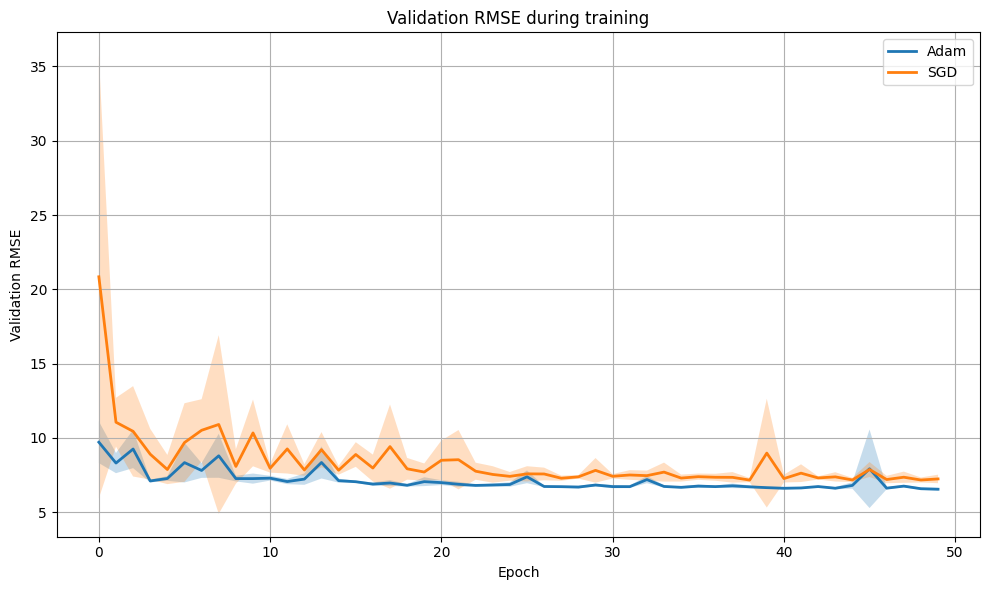}
    \caption{Comparison of the convergence behavior of SGD and Adam on WP-1. Curves show the average validation RMSE on the held-out 2024–2025 evaluation set over 5 independent runs, with shaded regions indicating one standard deviation.}
    \label{optimizer-comparison}
\end{figure}

To determine whether these optimization-related factors influenced the conclusions drawn from V1 and V2, we revisited the optimization strategy for the V3 dataset. Specifically, we replaced SGD with the Adam optimizer, which combines adaptive learning rates with momentum-based updates to improve convergence. As illustrated in 
Figure~\ref{optimizer-comparison}, Adam converges faster and more stably than SGD across the evaluated experiments.

Based on these findings, we performed all subsequent experiments in this paper using Adam. Using a single optimization strategy lets the remainder of the study focus on the effects of the proposed architectures and experimental configurations while minimizing variability introduced by the optimization process. Therefore, unless otherwise stated, all results presented in the following sections correspond to models trained with the Adam optimizer. Although Adam generally converges well before 50 epochs, as shown in both the experimental results and Figure~\ref{optimizer-comparison}, we keep this training duration to ensure a consistent evaluation protocol and consistent comparisons across the five independent runs. Hence, all RMSE values reported below are those at the end of 50 epochs, not the best value within the 50 epochs, except in Appendix \ref{app:r2}.

\subsection{Benchmarking Single-Channel Models}
\label{sec:single-benchmark}

For single-channel models, we evaluated the Digital Typhoon Dataset V3 for pressure and wind regression tasks. Table~\ref{tab:single-setting} summarizes the experimental setting. 

\begin{table}[t]
\caption{Experimental settings for benchmarking single-channel models.}
\label{tab:single-setting}
\centering
\begin{tabularx}{\textwidth}{lX}
\toprule
Hyper-parameters & Values \\
\midrule
Base model & ResNet-18, ResNet-50 \\
Pretrained weights & ImageNet \\
Image type & Full ($512\times 512$), Resized ($224\times 224$), Cropped ($224\times 224$) \\
Data split & SS-L2 \\
Optimizer & Adam, lr 1e-4 (constant) \\
Batch size & 16 \\
Epochs & 50 \\
Input scaling & clipped to [170, 300] K and scaled to [0, 1] \\
\bottomrule
\end{tabularx}
\end{table}

\subsubsection{Pressure Regression Task}

Table~\ref{tab:north-results-pressure} and 
Table~\ref{tab:south-results-pressure} show the results for single-channel pressure regression experiments.

\begin{table}[t]
    \caption{RMSE (hPa) for the pressure regression task for the WP dataset for four channels and three image types.}
    \label{tab:north-results-pressure}
    \centering
    \begin{tabularx}{\textwidth}{llXXX} 
    \toprule
        Ch. & Model & Full & Resized & Cropped \\ 
      \midrule
        1 & ResNet-18 & \textbf{6.5618} (±0.06) & 6.9591 (±0.09) & 6.9054 (±0.05) \\
       1 & ResNet-50 & 7.3367 (±1.13) & 7.0446 (±0.22) & 8.0222 (±2.53) \\
          \midrule
        2 & ResNet-18 & \textbf{6.7844} (±0.09) & 7.2420 (±0.06) & 7.0621 (±0.09) \\
        2 & ResNet-50 & 8.1232 (±1.29) & 8.0211 (±0.43) & 7.0262 (±0.26) \\
        \midrule
  3 & ResNet-18 & \textbf{6.9385} (±0.05) & 7.4771 (±0.15) & 7.3650 (±0.10) \\
        3 & ResNet-50 & 7.1517 (±0.24) & 7.8674 (±0.21) & 7.3509 (±0.19) \\
    \midrule
       4 & ResNet-18 & \textbf{7.3386} (±0.08) & 7.9457 (±0.15) & 7.6114 (±0.05) \\
        4 & ResNet-50 & 8.7589 (±2.00) & 8.6971 (±0.22) & 7.6587 (±0.14) \\
         \bottomrule    
    \end{tabularx}
\end{table}
\begin{table}[t]
    \caption{The RMSE (hPa) of the pressure regression task for the AU dataset for four channels and three image types.}
    \label{tab:south-results-pressure}
    \centering
    \begin{tabularx}{\textwidth}{llXXX} 
    \toprule
        Ch. & Model & Full & Resized & Cropped \\ 
      \midrule
        1 & ResNet-18 & \textbf{7.2015} (±0.12) & 7.6933 (±0.15) & 7.3887 (±0.31) \\
       1 & ResNet-50 & 7.3342 (±0.27) & 8.2380 (±0.35) & 7.5673 (±0.39) \\
          \midrule
        2 & ResNet-18 & 7.2708 (±0.29) & 7.8543 (±0.51) & \textbf{7.2427} (±0.12) \\
        2 & ResNet-50 & 8.5820 (±0.70) & 8.1601 (±0.42) & 8.2518 (±0.68) \\
        \midrule
  3 & ResNet-18 & 8.5809 (±1.36) & \textbf{7.9710} (±0.27) & 8.7664 (±1.56) \\
        3 & ResNet-50 & 9.5891 (±1.46) & 9.0027 (±0.71) & 8.4390 (±0.63) \\
    \midrule
       4 & ResNet-18 & 9.7814 (±0.31) & 10.2822 (±0.15) & \textbf{9.3286} (±0.17) \\
        4 & ResNet-50 & 10.2868 (±0.55) & 11.9096 (±1.75) & 9.5803 (±0.33) \\
         \bottomrule    
    \end{tabularx}
\end{table}

For the WP dataset, ResNet-18 yields the best results. The lowest error is observed for Channel 1 using "Full" images, where the model achieves an RMSE of 6.56 ± 0.06 hPa. More generally, ResNet-18 yields relatively consistent performance across all channels, with error increasing gradually as the channel number increases. Conversely, ResNet-50 shows no improvement despite its greater depth. Its performance is generally inferior to ResNet-18 and more variable, as evidenced by high standard deviations in certain configurations (e.g., 8.02 ± 2.53 hPa for Channel 1 in "Cropped" mode or 8.76 ± 2.00 hPa for Channel 4 in "Full" mode). This variability suggests that the model is less stable during training. Regarding image types, "Full" images yield the best results in most cases. "Resized" versions consistently degrade performance, likely because resizing loses detail. "Cropped" images sometimes offer a slight improvement over resized images, particularly with ResNet-50 on channels 2 and 4, but "Full" images generally outperform them. 

The results for the AU dataset generally follow the same trends as on WP, though with higher error rates. The best configuration is again ResNet-18 using Channel 1 and "Full" images, achieving an RMSE of 7.20 ± 0.12 hPa. As before, ResNet-18 generally outperforms ResNet-50. The performance gap between the two architectures is even more pronounced on this dataset; ResNet-50 yields RMSE values exceeding 10 hPa in certain configurations—specifically for Channel 4 with "Full" and "Resized" images. 

Image type also influences performance. "Full" images remain the most effective in most cases. "Resized" images generally produce the highest errors, whereas "Cropped" images sometimes recover some lost performance, though they do not surpass the results achieved with "Full" images. The spectral channel also matters: higher-numbered channels generally show higher RMSE, with values exceeding 9 hPa. See also Appendix~\ref{app:r2} for the epoch-to-epoch fluctuation on AU.

\subsubsection{Wind Regression Task}

Table~\ref{tab:north-results-wind} and Table~\ref{tab:south-results-wind} show the results for single-channel wind regression experiments.

\begin{table}[t]
    \caption{RMSE (kt) for the wind regression task for the WP dataset for four channels and three image types.}
    \label{tab:north-results-wind}
    \centering
    \begin{tabularx}{\textwidth}{llXXX} 
    \toprule
        Ch. & Model & Full & Resized & Cropped \\ 
      \midrule
        1 & ResNet-18 & \textbf{15.7493} (±1.09) & 16.9485 (±0.41) & 16.1151 (±0.27) \\
        1 & ResNet-50 & 15.9613 (±0.32) & 15.9931 (±0.06) & 15.8842 (±0.87) \\
      \midrule
        2 & ResNet-18 & \textbf{15.6748} (±0.80) & 17.0907 (±0.46) & 16.0942 (±0.28) \\
        2 & ResNet-50 & 15.9355 (±0.33) & 17.1484 (±1.00) & 16.1344 (±0.18) \\
      \midrule  
      3 & ResNet-18 & \textbf{16.0743} (±0.67) & 17.4122 (±0.66) & 16.3196 (±0.15) \\
        3 & ResNet-50 & 16.1426 (±0.40) & 16.8639 (±0.24) & 16.4675 (±0.60) \\
      \midrule
      4 & ResNet-18 & \textbf{16.0726} (±0.85) & 18.3602 (±0.32) & 16.7479 (±0.31) \\
        4 & ResNet-50 & 16.6024 (±0.32) & 18.1410 (±0.41) & 16.8533 (±0.26) \\
           \bottomrule
    \end{tabularx}
\end{table}
\begin{table}[t]
    \caption{RMSE (kt) for the wind regression task for the AU dataset for four channels and three image types.}
    \label{tab:south-results-wind}
    \centering
    \begin{tabularx}{\textwidth}{llXXX} 
    \toprule
        Ch. & Model & Full & Resized & Cropped \\ 
      \midrule
        1 & ResNet-18 & 11.4344 (±0.13) & 11.6104 (±0.17) & 11.5828 (±0.24) \\
        1 & ResNet-50 & \textbf{10.7545} (±0.17) & 11.9051 (±0.12) & 11.0413 (±0.86) \\
      \midrule
        2 & ResNet-18 & 13.2135 (±0.33) & 13.4230 (±0.18) & \textbf{12.6630} (±0.19) \\
        2 & ResNet-50 & 13.6150 (±0.42) & 13.7778 (±0.54) & 13.7715 (±0.63) \\
      \midrule  
      3 & ResNet-18 & 12.9419 (±0.54) & 13.2351 (±0.74) & 12.7642 (±0.44) \\
        3 & ResNet-50 & \textbf{12.5016} (±0.31) & 13.1745 (±0.52) & 13.1644 (±0.50) \\
      \midrule
      4 & ResNet-18 & \textbf{14.0658} (±0.19) & 15.0909 (±0.20) & 15.1175 (±0.14) \\
        4 & ResNet-50 & 14.9084 (±0.17) & 14.5259 (±0.34) & 14.8645 (±0.54) \\     
           \bottomrule
    \end{tabularx}
\end{table}

For the WP dataset, ResNet-18 with Channel 2 and "Full" images yields the lowest RMSE, 15.67 $\pm$ 0.80 kt. Channel 1 with "Full" images achieves a very similar RMSE of 15.75 $\pm$ 1.09 kt. Given the overlapping variability across runs, these results do not provide strong evidence that Channel 2 is systematically superior to Channel 1 for wind regression. Overall, ResNet-18 performs competitively across the evaluated channels, while error tends to increase for the higher-numbered channels. In contrast, ResNet-50 does not consistently improve over ResNet-18 despite its greater depth. Its performance is generally comparable to or worse than ResNet-18 and shows higher variability across configurations. Regarding image types, "Full" images yield the best results in most cases. "Resized" versions consistently degrade performance, likely because resizing loses detail. "Cropped" images sometimes offer a slight improvement over resized images, particularly with ResNet-50 on channels 2 and 4, but "Full" images generally outperform them. Finally, a clear trend emerges regarding the channels: performance generally declines from Channel 1 to Channel 4.

The results for the AU dataset generally follow the same trends as those observed for WP, albeit with some differences. The best configuration for ResNet-18 is Channel 1 with "Full" images, achieving an RMSE of 11.43 ± 0.13 kt, while ResNet-50 performs best with Channel 1 and "Full" images, achieving an RMSE of 10.75 ± 0.17 kt. Unlike the WP dataset, ResNet-50 occasionally outperforms ResNet-18 in specific configurations, such as for Channel 1 with "Full" images and Channel 3 with "Full" images (12.50 ± 0.31 kt). The influence of image type also follows the same trends as WP.

\subsubsection{Comparison between Single-Channel Models}
\label{subsec:single-channel-comparison}
A comparison between the WP and AU datasets reveals task-dependent performance asymmetries. For pressure regression, WP consistently achieves lower RMSE values than AU across the evaluated channels and image types. For example, WP achieves a best pressure RMSE of 6.56 hPa (Channel 1, Full images, ResNet-18), compared with 7.20 hPa for AU under the same configuration. 

In contrast, AU substantially outperforms WP on the wind regression task. The lowest RMSE observed in WP is 15.67 kt (Channel 2, Full images, ResNet-18), whereas the lowest RMSE observed in AU is 10.75 kt (Channel 1, Full images, ResNet-50). Thus, the AU dataset achieves a lower wind regression error by approximately 4.9 kt in the best-performing configurations. 

These results show that benchmark performance differs systematically between the WP and AU datasets by target variable. WP shows lower RMSE for pressure regression, whereas AU shows lower RMSE for wind regression. This asymmetry stems from several factors across the two hemispheres.
\begin{enumerate}
\item The WP dataset is larger than the AU dataset, which favors the WP models for pressure regression.
\item Superior wind-regression performance in the AU dataset may reflect differences in data definitions and quality between hemispheres, driven by different operational procedures in regional meteorological agencies \cite{agencies_diff}. But it is unclear how this affects model performance.
\end{enumerate}

This paper does not study the cause of performance gaps between hemispheres, so we leave this problem for future work. 

\subsection{Benchmarking Multi-channel Models}
\label{sec:multi-benchmark}

\begin{table}[t]
    \centering
     \caption{Results of the pressure regression task for the WP dataset, comparing single-channel and multi-channel inputs. Results from the single-channel dataset with the same configuration are taken from Table~\ref{tab:north-results-pressure}.}
    \label{tab:wp-single-vs-multi-pressure}
    \begin{tabular}{lll}
        \toprule
        Input channels & RMSE (hPa) & Single-channel dataset\\
        \midrule
        Channel 1 & 6.9270 (±0.12) & 6.5618 (±0.06) \\
        Channel 2 & 6.9605 (±0.07) & 6.7844 (±0.09)\\
        Channel 3 & 7.3958 (±0.58) & 6.9385 (±0.05)\\
        Channel 4 & 7.3973 (±0.17) & 7.3386 (±0.08)\\
        \midrule
        Channel 1 + 2 & 6.9265 (±0.11) & -\\
        Channel 1 + 3 & 7.1512 (±0.24) & -\\
        Channel 1 + 2 + 3 & \textbf{6.8308} (±0.10) & - \\
        \bottomrule
    \end{tabular}
\end{table}

\begin{table}[t]
    \centering
    \caption{Results of the pressure regression task for the AU dataset, comparing single-channel and multi-channel inputs. Results from the single-channel dataset with the same configuration are taken from Table~\ref{tab:south-results-pressure}.}
    \label{tab:au-single-vs-multi-pressure}
    \begin{tabular}{lll}
        \toprule
        Input channels & RMSE (hPa) & Single-channel dataset \\
        \midrule
        Channel 1 & 8.2839 (±0.33) & 7.2015 (±0.12)\\
        Channel 2 & \textbf{8.2370} (±0.62) & 7.2708 (±0.29)\\
        Channel 3 & 8.4316 (±0.60) & 8.5809 (±1.36)\\
        Channel 4 & 9.7118 (±0.27) & 9.7814 (±0.31)\\
        \midrule
        Channel 1 + 2 & 8.4443 (±0.10) & - \\
        Channel 1 + 3 & 8.4714 (±0.42) & - \\
        Channel 1 + 2 + 3 & 8.5677 (±0.22) & -\\
        \bottomrule 
    \end{tabular}
\end{table}

For multi-channel benchmarking, we want to assess the effect of adding new channels. 
For a fair comparison, we must use the same dataset to train and evaluate single-channel and multi-channel models. The multi-channel dataset, introduced in Section~\ref{sec:dataset-definitions}, addresses this by creating a filtered dataset that is the intersection of the four channels. If we train and evaluate all models on this dataset, we can train single-channel and multi-channel models on the same dataset for both WP and AU hemispheres and directly compare channel relevance on the same basis. However, the multi-channel dataset has smaller temporal coverage and a smaller dataset size due to filtering. 
Here, single-channel baselines in Table~\ref{tab:north-results-pressure} and Table~\ref{tab:south-results-pressure} show the effect of this filtering. This compares the same model trained on a larger dataset (single-channel) and a smaller dataset (multi-channel), so performance differences can be attributed to training-set size. 

We used ResNet-18 and "Full" as the image type, as this setting performed best in the single-channel benchmark (Section~\ref{sec:single-benchmark}). For data splitting, we employed season-level splitting (SS-L2), with the 2024--2025 seasons serving as the evaluation set. 

Table~\ref{tab:wp-single-vs-multi-pressure} and Table~\ref{tab:au-single-vs-multi-pressure} summarize performance on the pressure regression task across different combinations of input channels, along with a comparison between single-channel and multi-channel datasets for the same configuration.

We first focus on single-channel models on multi-channel datasets. Channel 1 and Channel 2 performed best
on both datasets with nearly identical RMSE (6.93 and 6.96 hPa on WP, 8.28 and 8.24 hPa on AU), a difference
well within the variability across runs, while Channel 4 performed worst on AU and, together with
Channel 3, on WP. This means that the two infrared channels, Channel 1 and Channel 2, contain the most
valuable information for pressure regression, while the other channels are less useful. Note that this is
not the same argument as in Section~\ref{sec:single-benchmark}, because the single-channel results are compared on different
dataset sizes across channels. We compare multi-channel results on the same dataset size; still,
Channel 1 performed among the best.

Based on this result, we tested channel combinations for multi-channel benchmarking. We always include Channel 1 and add Channels 2 and 3 to test all possible combinations. As a result, we test Channel 1 + 2, Channel 1 + 3, and Channel 1 + 2 + 3. We exclude Channel 4 from these combinations because of its distinct spectral and diurnal characteristics, as discussed in Section~\ref{sec-dataset}. 

On WP, Channel 1 + 2 + 3 achieved the lowest RMSE among all input configurations (6.8308 hPa),
0.10 hPa lower than the Channel 1 model on the same dataset (6.9270 hPa); this difference is comparable to
the variability across runs. On AU, the multi-channel models did not outperform the single-channel models:
Channel 1 + 2 achieved the lowest RMSE among the multi-channel models (8.4443 hPa), which is higher than
Channel 1 (8.2839 hPa) and Channel 2 (8.2370 hPa). Overall, the results indicate comparable performance
rather than a clear advantage of multi-channel inputs.

More importantly, both tables show that, despite the intuitive appeal of incorporating multiple input channels to improve model performance, the empirical evidence does not consistently justify the additional computational cost. Across both datasets, the performance gains from combining channels are marginal. 

\subsection{Trade-off between More Spectral Data and Smaller Dataset Size}

We again compare the single-channel regression task performed on the original single-channel datasets with the same task performed on the multi-channel intersection dataset (Section~\ref{sec:dataset-definitions}), using a single channel as model input. 
The learning objective and input channel are identical within each comparison, while the underlying datasets differ in temporal coverage and number of available observations. The multi-channel dataset contains 96,579 observations for WP and 40,236 for AU, mainly limited by Channel 4 availability. 

Table~\ref{tab:north-results-pressure} and Table~\ref{tab:south-results-pressure} show that the best single-channel results obtained on the original datasets are lower than those obtained when the same single-channel inputs are evaluated on the multi-channel intersection dataset 
(Table~\ref{tab:wp-single-vs-multi-pressure} and Table~\ref{tab:au-single-vs-multi-pressure}). 
For the WP dataset, the RMSE for Channel~1 increases from 6.56~hPa to 6.93~hPa, while for the AU dataset it increases from 7.20~hPa to 8.28~hPa. 
Similar degradations are observed for Channel 2 on both datasets and for
Channel 3 on WP, whereas the differences for Channel 4 and for Channel 3 on AU are within the variability
across runs.

The observed performance gap coincides with the construction of the multi-channel dataset. 
An observation is retained in the intersection dataset only when all four spectral channels are simultaneously available, whereas an original single-channel dataset can retain observations even when observations from other channels are unavailable. As a consequence, the multi-channel dataset has shorter temporal coverage and substantially fewer observations than the corresponding original single-channel datasets, reducing the amount of training data available to the model. 

For example, for WP-1, the multi-channel dataset contains 96,579 observations with 6.93 hPa RMSE,
while the single-channel dataset contains 202,986 observations with 6.56 hPa RMSE. For WP-4, the
multi-channel dataset contains 96,579 observations with 7.40 hPa RMSE, while the single-channel
dataset contains 96,611 observations with 7.34 hPa RMSE.
Channel 4's performance differs little between the single-channel and multi-channel datasets because it is the least affected channel. The multi-channel dataset is constrained by the channel with the shortest temporal coverage, Channel 4, so the two datasets are nearly identical.  

These results highlight a trade-off between spectral completeness and dataset coverage. The multi-channel dataset enables experiments using complementary spectral information, but its construction restricts observations to periods when all four channels are simultaneously available. The resulting reduction in dataset size and temporal coverage must therefore be considered when interpreting the multi-channel benchmark results.

Overall, the results can be summarized as follows.
\begin{enumerate}
\item Channel 1, together with the similar infrared Channel 2, contains more relevant information than the other channels.
\item On WP, Channel 1 + 2 + 3 performed comparably to Channel 1, but adding other channels to
 Channel 1 does not clearly or consistently improve performance on either dataset.
\item To use multi-channel models, we need to reduce the dataset size because multiple channels are not always available. This reduction in dataset size cannot be compensated by the advantage of multi-channel models. 
\end{enumerate}

However, these results should not be interpreted as evidence that the additional spectral information is intrinsically uninformative. For example, the difference between Channel 1 and Channel 2 is used in the split-window method and is known to be informative for some cloud classification tasks, but we did not explore combining multiple channels using arithmetic operations. 

\subsection{Key Benchmarking Findings}
\label{subsec:key-benchmarking-findings}

The main findings of the benchmarking experiments can be summarized as follows:

\begin{itemize}
    \item \textbf{Optimizer:} Adam converges faster and more stably than SGD and is therefore used for all subsequent experiments.

    \item \textbf{Single-channel pressure regression:} ResNet-18 generally provides better, more stable performance than ResNet-50 across the evaluated configurations, although ResNet-50 achieves a slightly lower RMSE in some individual settings. Channel 1 with Full-resolution images provides the best performance, achieving an RMSE of 6.56 hPa on WP and 7.20 hPa on AU.

    \item \textbf{Single-channel wind regression:} ResNet-18 with Full-resolution images achieves the best observed performance for WP using Channel 2 (15.67 kt). For AU, ResNet-50 with Channel 1 and Full-resolution images achieves the best result (10.75 kt).

    \item \textbf{Hemispheric differences:} Performance differs between the WP and AU datasets by target variable. WP exhibits lower RMSE for pressure regression, whereas AU exhibits substantially lower RMSE for wind regression. These differences reflect dataset- and task-specific characteristics rather than experimental choices.

    \item \textbf{Multi-channel inputs:} Adding multiple spectral channels does not clearly improve the best single-channel model for pressure regression. 
    The best multi-channel models achieve an RMSE of 6.83 hPa on WP (Channel 1 + 2 + 3) and 8.44 hPa on AU (Channel 1 + 2).
    
    \item \textbf{Dataset size versus spectral information:} The multi-channel dataset is smaller because we retain an observation only when all required channels are simultaneously available. The resulting reduction in dataset size and temporal coverage could not compensate for the advantage of combining multiple channels. 

    \item \textbf{Overall conclusion:} Under the current experimental setting, Channel 1 with Full-resolution images and ResNet-18 provides the strongest and most consistent single-channel configuration for pressure regression, while the additional spectral information available in the multi-channel dataset does not translate into a clear performance advantage.
\end{itemize}

\section{Analyzing Systematic Bias in Intensity Regression}
\label{sec:improving-extreme}

\subsection{Systematic Bias in Intensity Regression}

As discussed in Section~\ref{sec:usecases}, dataset users have already noticed degraded performance in intensity regression for strong typhoons. 
We did not analyze performance by intensity in our previous benchmarking for the Digital Typhoon Dataset because we averaged regression performance over all typhoon strengths, including strong and weak typhoons. By averaging, we could compare model configurations, channels, and hemispheres. Breaking down performance by intensity may provide practical guidance for improving the regression model. 

The first approach is to employ more complex model architectures, such as ViT and ConvNeXT. However, the previous result showed that a simpler model, ResNet-18, generally performs better than a more complex model, ResNet-50, suggesting that increasing model complexity may not be the right path to improving intensity regression. 

To explore an alternative path, we examine the error distribution across intensities, with a special focus on strong typhoons. This is because strong typhoons have a larger impact on society. For disaster preparedness, the loss function for forecasting error is not uniform, and performance improvements for stronger typhoons matter more than for weaker ones. However, when we evaluate performance on the entire dataset, it tends to optimize for weak typhoons because the dataset is biased toward them.  Thus, this study aims to compensate for this bias and improve performance for strong typhoons.

Here we distinguish three possible sources of error: data sparsity for strong typhoons, regression to the mean induced by the MSE objective, and limitations in the information available from a single satellite image. These effects are not mutually exclusive. The first two may reflect the training strategy, whereas the latter may reflect an intrinsic limitation of the input.

In this section, we use the term systematic bias descriptively to denote the dependence of prediction error on the true pressure. For a pressure bin $B$, we define the bias as the conditional mean error:
\begin{equation}
b(B) = \mathbb{E}[\hat{y} - y \mid y \in B],
\end{equation}
where $\hat{y}$ denotes the predicted pressure and $y$ the true central pressure. A non-zero value of $b(B)$ does not by itself imply a defect in the regression model. In particular, when the target cannot be fully determined from the input image, an MSE-trained model may exhibit regression-to-the-mean behavior, producing predictions closer to the average pressure. We therefore interpret the observed intensity-dependent error as a descriptive property of the model predictions rather than attributing it to a single underlying cause.

We evaluate the bias between predicted and true pressure values using a sliding-window validation approach. Specifically, we trained the WP-1 dataset using ResNet-18 with full-resolution images for each consecutive-year pair from 1978 to 2024, excluding the target pair from the training set, and validated the model on that pair alone. This resulted in 47 overlapping temporal validation splits (e.g., 1978–1979, 1979–1980, ..., 2024–2025), each assessing the model's ability to generalize to unseen temporal data, assuming no causal relationship across seasons. This analysis uses the 47 temporal splits directly rather than repeated runs on a fixed partition.

\begin{figure}[t]
    \centering
    \includegraphics[width=0.9\linewidth]{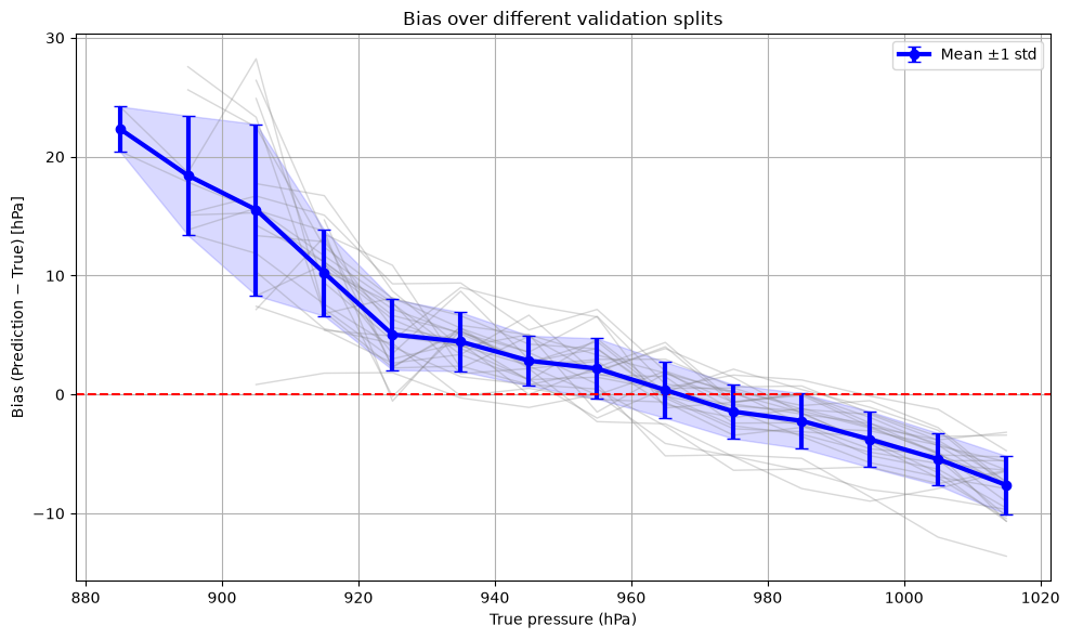}
    \caption{Systematic bias (Prediction - True) across seasonal splits for the pressure regression task (WP-1 dataset, ResNet-18, Channel 1, Full images). Gray curves represent individual splits (each corresponding to a consecutive-year pair from 1978–1979 to 2024–2025), while the blue curve shows the mean bias across all 47 overlapping temporal splits. The shaded area represents ±1 standard deviation.}    
    \label{fig:bias-validation}
\end{figure}

We evaluate performance on a pressure regression task, where lower central pressure means stronger typhoons. As a result, we find a systematic bias in intensity regression, as shown in Figure~\ref{fig:bias-validation}. The model underestimates strong typhoons and overestimates weak ones. Errors are especially large for pressures below 920 hPa across all splits, with the mean bias (blue curve) exceeding 10 hPa in the most extreme cases. This pattern is particularly pronounced for the strongest typhoons, for which the observed prediction error is substantially larger.
This may be related to the underrepresentation of strong typhoons in the training data, as less than 1\% of samples have pressures below 920 hPa. This result may reflect regression-to-the-mean behavior, where predictions for high-magnitude events are pulled toward the average.

To address systematic bias for strong typhoons, we explore two strategies in Section~\ref{subsec:weighting} and Section~\ref{subsec:balanced-mse}: sample weighting to account for the underrepresentation of extreme events, and an inverse-density weighting strategy inspired by BalancedMSE to further investigate the effect of target-density-aware weighting. These approaches aim to reduce the bias for extreme cases while maintaining robust performance across all pressure ranges.

\subsection{Sample Weighting}
\label{subsec:weighting}
To address the underrepresentation of strong typhoons, we implemented a multi-faceted weighting strategy inspired by imbalanced learning techniques. This approach combines static and dynamic components to emphasize rare or hard-to-predict samples. All experiments in this section use the WP-1 dataset with the SS-L2 split (2024--2025), retaining the last two seasons for evaluation, as described in Table~\ref{tab:data-split-wp}. Each weighting configuration was evaluated using five independent training runs on this fixed partition. For the adaptive weighting procedure, we compute the model's per-bin RMSE on the training set at the end of each training epoch and use the resulting values to update the weights for the subsequent epoch. 
We reserve the held-out 2024–2025 set exclusively for reporting final benchmark performance and do not use it to compute the per-bin RMSE that drives adaptive weight updates. This protocol is intentionally used to study whether dynamically emphasizing poorly predicted pressure ranges can reduce the systematic bias toward the mean.

For the weighting mechanism, we first discretize the pressure range (850–1050 hPa) into 20 equal-width bins (10 hPa each). This binning is computed as follows:
For a pressure value $p$, its corresponding bin index $b$ is determined by:
\begin{equation}
b = \left\lfloor \frac{p - 850}{10} \right\rfloor,
\end{equation}
where $\left\lfloor \cdot \right\rfloor$ denotes the floor function.
This ensures that each bin covers a 10 hPa interval (e.g., bin 0: [850, 860), bin 1: [860, 870), ..., bin 19: [1040, 1050]). We selected the minimum and maximum pressure values not as the dataset's actual extremes, but with a margin. This binning allows us to assign weights based on the frequency of samples in each pressure interval, ensuring that rare events (e.g., pressures below 920 hPa) receive higher attention during training.

First, we assigned initial weights inversely proportional to the sample frequency in each pressure bin. For a bin $i$ with frequency $f_i$, the initial weight is defined as:
\begin{equation}
w_i^{(0)} = \frac{1}{f_i}.
\end{equation}
These weights were then normalized to ensure their mean equals 1, preventing excessive bias toward rare classes:
\begin{equation}
w_i^{(0)} \leftarrow \frac{w_i^{(0)}}{\text{mean}(\mathbf{w}^{(0)})}.
\end{equation}
This static weighting directly targets the class imbalance by emphasizing bins with fewer samples, such as those corresponding to low pressures below 920 hPa.

However, to further refine the model's focus on challenging cases, we introduced an adaptive weighting mechanism. At the end of each training epoch, we compute the model's per-bin RMSE on the training set. We then use this per-bin RMSE to update the weights for the next epoch. Specifically, we used a smoothed update rule:
\begin{equation}
w_i^{(t+1)} = (1 - \alpha) \cdot w_i^{(t)} + \alpha \cdot \left( w_i^{(0)} \cdot \frac{\text{RMSE}_i}{\overline{\text{RMSE}}} \right),
\end{equation}
where $\alpha = 0.2$ is a smoothing factor, $\text{RMSE}_i$ is the Root Mean Squared Error for bin $i$ computed on the training set, and $\overline{\text{RMSE}}$ is the mean RMSE across all bins. 
This ensures that bins with higher prediction errors receive progressively more attention. After each update, we renormalized the weights to maintain a mean of 1. The held-out 2024--2025 seasons therefore never influence the weights and are reserved for performance evaluation on unseen seasons.

We applied the weights in two complementary ways. First, we used a weighted loss, scaling the MSE loss for each sample by its corresponding bin weight. For a batch of $N$ samples, this weighted loss is computed as:
\begin{equation}
\mathcal{L} = \frac{1}{N} \sum_{j=1}^N w_{b_j} \cdot (y_j - \hat{y}_j)^2,
\end{equation}
where $w_{b_j}$ is the weight for the bin $b_j$ of sample $j$, $y_j$ is the true pressure, and $\hat{y}_j$ is the predicted pressure. This forces the model to prioritize under-represented or poorly predicted bins. Additionally, we implemented weighted sampling using a \texttt{WeightedRandomSampler} in the training data loader, which increased the probability of selecting samples from bins with higher weights. This further emphasizes rare or hard-to-predict cases during training.

For all weighting schemes, we used the same weighting factors for both the weighted loss and the WeightedRandomSampler. Thus, the reported results combine loss weighting and weighted sampling, rather than evaluating these two mechanisms independently.

\begin{figure}[t]
    \centering
    \subfloat[Mean prediction error with $1/f$ weighting (RMSE = 8.59).]{
        \includegraphics[height=4.2cm]{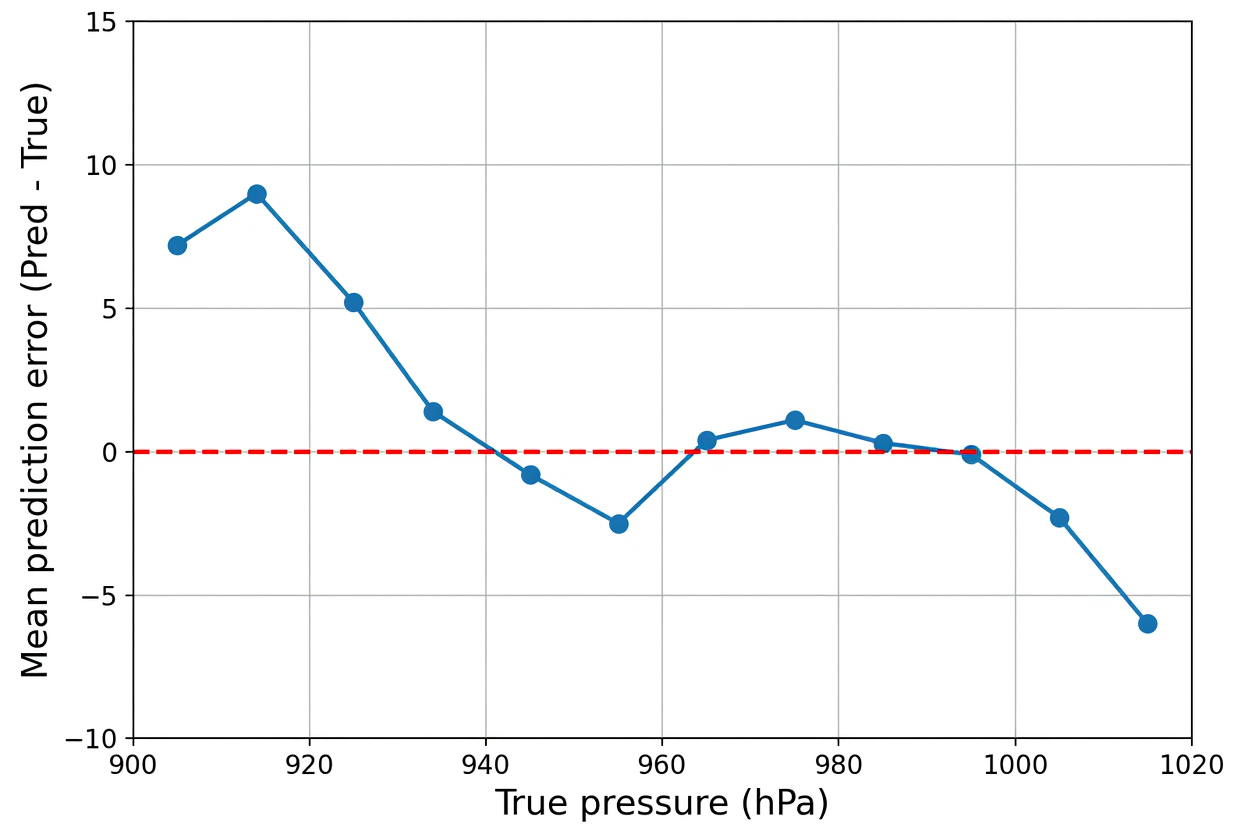}
        \label{fig:weighting-1f}
    }
    \hfill
    \subfloat[Mean prediction error with $1/f^{0.25}$ weighting (RMSE = 7.04).]{
        \includegraphics[height=4.2cm]{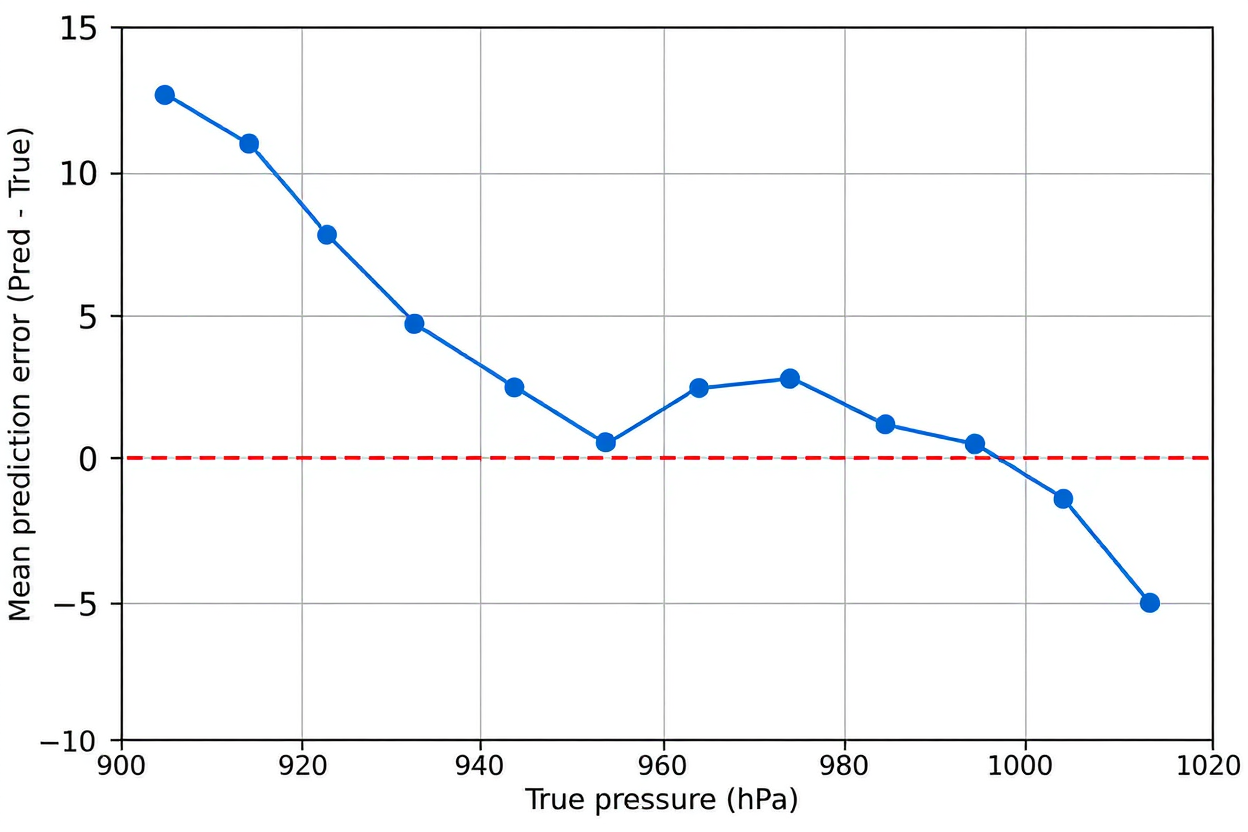}
        \label{fig:weighting-f025}
    }
    \caption{Trade-off between bias reduction for strong typhoons and global RMSE for different weighting schemes for the WP-1 dataset. Results are averaged over five independent training runs using the same fixed training protocol. The red dashed line indicates zero error.}
    \label{fig:weighting-tradeoff}
\end{figure}

To evaluate the impact of weighting intensity, we show the results of two schemes, $1/f^{0.25}$ and $1/f$ in 
Figure~\ref{fig:weighting-1f} and Figure~\ref{fig:weighting-f025}.
These figures show that stronger weightings, such as $1/f$ (Figure~\ref{fig:weighting-1f}, RMSE = 8.59), reduce bias for extreme cases but increase global RMSE. In contrast, the $1/f^{0.25}$ scheme (Figure~\ref{fig:weighting-f025}, RMSE = 7.04) preserves overall RMSE while moderately improving predictions of rare pressures. This trade-off shows that stronger weightings reduce bias for strong typhoons but at the expense of global performance.

Even with stronger weightings, the bias curve shape remains similar (Figure~\ref{fig:weighting-tradeoff}). While the bias magnitude decreases, the model still exhibits an intensity-dependent prediction error, with predictions for low pressures remaining closer to the average pressure. These results show that target weighting can mitigate the magnitude of the
observed bias for rare pressure ranges, but does not by itself establish the underlying cause of the remaining intensity-dependent pattern. This motivates the exploration of complementary approaches, such as loss functions tailored for imbalanced regression (e.g., \textit{BalancedMSE}) or prediction
calibration techniques.

\subsection{Inverse-Density Weighting Inspired by BalancedMSE}
\label{subsec:balanced-mse}
The experiments in this section are conducted for the WP-1 dataset with the same SS-L2 split (2024--2025) as in Section~\ref{subsec:weighting}, using the same 10 hPa pressure bins. The standard Mean Squared Error (MSE) loss is dominated by the most frequent pressure ranges, which may contribute to the systematic regression-to-the-mean behavior observed for strong typhoons. This issue is well documented in imbalanced regression, where the distribution of training targets can strongly affect regression performance \cite{ren2022balanced}.

The original BalancedMSE formulation proposed by Ren et al.~\cite{ren2022balanced} is a distribution-aware probabilistic reformulation of the regression objective. It is therefore conceptually different from simply multiplying the per-sample MSE by an inverse-frequency weight. Our implementation does not reproduce the full probabilistic formulation of the original method. Instead, we investigate a simplified inverse-density weighting strategy inspired by BalancedMSE, in which we adjust each sample's contribution according to the empirical density of its target bin.

For a given sample $j$ belonging to bin $b_j$, the loss is scaled by the inverse of the bin's density $d(b_j)$:
\begin{equation}
\mathcal{L}_{\text{IDW}} =
\frac{1}{N}
\sum_{j=1}^{N}
\frac{(y_j-\hat{y}_j)^2}{d(b_j)},
\label{eq:inverse-density}
\end{equation}
where $y_j$ is the true pressure, $\hat{y}_j$ is the predicted pressure, and $d(b_j)$ denotes the empirical density of training samples in pressure bin $b_j$. In our implementation, we estimate the density from the empirical frequency of each pressure bin. Consequently, $1/d(b_j)$ is equivalent to inverse-frequency weighting up to the normalization used in the implementation.

This formulation should not be confused with the original BalancedMSE proposed by Ren et al.~\cite{ren2022balanced}. Our inverse-density formulation corresponds to the weighted-loss component of the \(1/f\) weighting scheme introduced in Section~\ref{subsec:weighting}, up to the normalization used in the implementation. However, the two experiments differ in their training procedures: the experiments in Section~\ref{subsec:weighting} combine weighted loss with a \texttt{WeightedRandomSampler}, whereas this experiment uses no weighted sampler. Therefore, the two experiments are not equivalent in implementation. Instead, this experiment isolates the effect of loss weighting, while Section~\ref{subsec:weighting} evaluates the combined effect of loss weighting and weighted sampling.

\begin{figure}[t]
    \centering
    \includegraphics[width=0.75\linewidth]{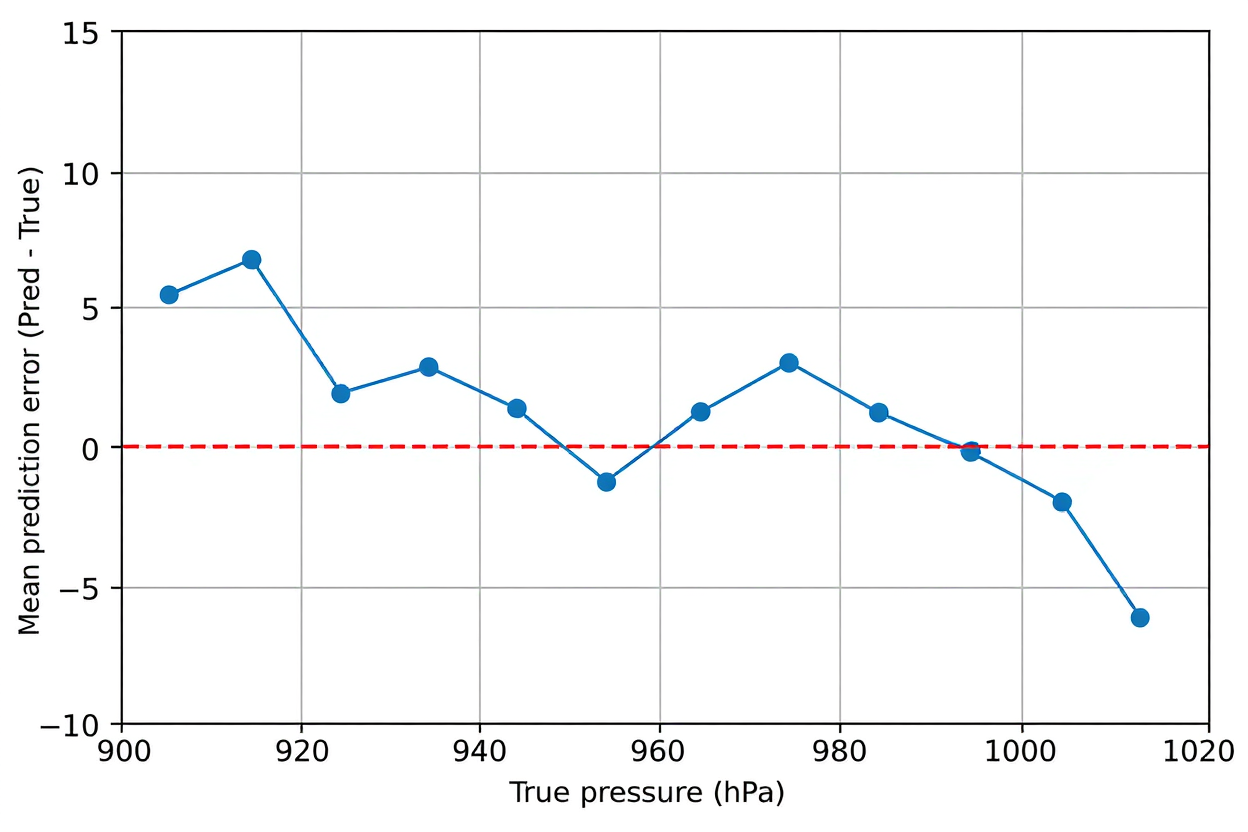}
    \caption{Mean prediction error (Pred - True) for the WP-1 dataset using inverse-density weighting (global RMSE = 8.42). Results are averaged over five independent training runs using the same fixed training protocol. The red dashed line indicates zero error.}
    \label{fig:balancedmse-curve}
\end{figure}

Figure~\ref{fig:balancedmse-curve} shows that inverse-density weighting reduces bias magnitudes at low pressures.

Although the values in Figure~\ref{fig:balancedmse-curve} are not directly comparable to those in Figure~\ref{fig:bias-validation} because of the different validation protocol, the overall shape of the bias curve is similar to Figures~\ref{fig:bias-validation} and \ref{fig:weighting-tradeoff}, while its magnitude is smaller than in Figure 4, particularly below 940 hPa.

Since this experiment applies the \(1/f\)-based weighting only through the loss, its results complement the combined loss-and-sampling experiment in Section~\ref{subsec:weighting}. The comparison therefore helps determine whether the observed bias reduction depends on the sampling strategy in addition to the loss weighting.

The result is consistent with BalancedMSE's motivation: reducing the influence of a highly imbalanced target distribution in regression. However, because the present implementation is equivalent to inverse-frequency weighting, we should not attribute the observed improvement to the probabilistic BalancedMSE formulation of Ren et al.~\cite{ren2022balanced}. Instead, it demonstrates the effect of a simpler density-aware weighting strategy in this typhoon intensity regression setting.

\subsection{Future Directions for Intensity Regression}
\label{subsec:future-work}

The results presented in this section suggest several directions for future research on intensity regression for strong typhoons. Since the Digital Typhoon Dataset V3 already incorporates all the historically available observations under its current construction, future improvements should not rely simply on increasing the number of observations. Instead, they should focus on using the existing data more effectively and developing models better suited to the characteristics of extreme events.

One promising direction is to investigate alternative learning strategies for imbalanced regression, including improved formulations of density-aware loss functions and adaptive weighting schemes. Such methods could reduce regression-to-the-mean behavior while limiting the overall performance degradation seen with stronger weighting schemes. Another direction is to investigate prediction calibration or uncertainty-aware approaches, which could provide more reliable estimates for the rarest and most intense typhoons.

Future work could also explore using additional information already associated with the observations, such as temporal context or complementary meteorological variables, provided these extensions remain consistent with the benchmark design. In particular, incorporating temporal sequences rather than treating satellite images independently may help models better capture the physical evolution of rapidly intensifying or weakening tropical cyclones.

Data augmentation requires particular caution for this dataset. Unlike generic image-recognition tasks, transformations applied to satellite observations may alter physically meaningful structures or introduce unrealistic cyclone configurations. Consequently, conventional geometric or photometric augmentation is inappropriate for this dataset, and this study intentionally did not use it. If future work considers augmentation, it should instead use transformations that preserve the physical interpretation of the observations and validate them carefully.

Overall, future research should focus on better exploiting existing observations, developing imbalance-aware and physically informed learning strategies, and better representing the temporal and physical characteristics of strong tropical cyclones.

\section{Conclusion}
In this paper, we presented the Digital Typhoon Dataset V3, extended it to spectral dimensions, and opened new research avenues for tropical cyclone analysis. The spectral dimension enables the development of multi-channel models, addressing a long-standing need in tropical cyclone research. However, our benchmarking experiments (Section~\ref{sec:multi-benchmark}) show that multi-channel inputs do not clearly improve pressure regression in the current setting. This highlights a critical trade-off: while richer spectral information is theoretically valuable, the reduction in dataset size (due to missing channels) may limit the model's ability to learn robust features. This limitation is particularly relevant for strong typhoons, which are already under-represented in the data.

Our experiments in Section~\ref{sec:improving-extreme} further investigate this challenge. The strategies we investigate—sample weighting and inverse-density weighting inspired by BalancedMSE—reduce bias for extreme cases but do not fully eliminate regression-to-the-mean behavior. This suggests that both dataset constraints and model limitations contribute to the difficulty of analyzing strong typhoons. 

Together, these contributions in dataset development and modeling strategies provide a foundation for future work to improve tropical cyclone intensity regression, particularly for extreme events.

\begin{ack}
Four of the authors, Victor Le Louarn, Noelia Falczuk, Ngoc Anh Tong, and Anja Veleva, have been supported by the international internship program of the National Institute of Informatics. Their affiliations are as of the time of the NII internship. 
\end{ack}

\bibliographystyle{plain}
\bibliography{references}

\clearpage
\appendix

\section{Comparison of V3-2025.R1 and V3-2025.R2}
\label{app:r2}

\begin{table}[t]
\caption{Number of TCs and images for the WP and AU datasets for V3-2025.R2. The only differences
from Table~\ref{tab:number-two-basins} (V3-2025.R1) are the AU images in Channels 1 and 2, which are 23 fewer
because of the removed GOES-9 images (Section~\ref{sec:correction}); the number of TCs is unchanged.}
\label{tab:dataset_r2}
\centering
    \begin{tabular}{lrrrr} 
\hline
Ch. & WP TCs & WP images & AU TCs & AU images \\
\hline
1 & 1,169 & 202,986 & 511 & 76,188 \\
2 &   744 & 142,583 & 339 & 58,842 \\
3 &   744 & 142,591 & 339 & 58,842 \\
4 &   502 &  96,611 & 223 & 40,238 \\
\hline
\end{tabular}
\end{table}

\begin{table}[t]
\caption{
Comparison of V3-2025.R1 and V3-2025.R2: mean ($\pm$ standard deviation) over five runs.
``Corrupted'' is the number (fraction) of training observations whose image was corrected in V3-2025.R2;
in addition, 23 training images were removed from each of AU-1 and AU-2. None of them are in the evaluation set. ``Last 10'' and ``Last 5'' are the medians of the
evaluation RMSE over epochs 41--50 and 46--50 of each run. All values are in hPa. A large AU-3 R2 value is due to 2 runs that ended on a spike.}
\label{tab:r2}
\centering
\small
\begin{tabular}{lllllll}
\hline
 & & & \multicolumn{3}{c}{Evaluation RMSE} & Training RMSE \\
Dataset & Corrupted & Release & Last 10 & Last 5 & Epoch 50 & Epoch 50 \\
\hline
WP-1 & 577 (0.30\%)   & R1 & 6.63 ($\pm$0.01) & 6.61 ($\pm$0.05) & 6.66 ($\pm$0.08) & 2.03 \\
     &                & R2 & 6.54 ($\pm$0.02) & 6.52 ($\pm$0.04) & 6.51 ($\pm$0.05) & 1.33 \\
WP-3 & 1,252 (0.94\%) & R1 & 7.10 ($\pm$0.08) & 7.09 ($\pm$0.09) & 7.07 ($\pm$0.11) & 2.89 \\
     &                & R2 & 6.98 ($\pm$0.05) & 7.04 ($\pm$0.10) & 7.05 ($\pm$0.23) & 1.34 \\
AU-1 & 39 (0.06\%)    & R1 & 7.53 ($\pm$0.19) & 7.60 ($\pm$0.23) & 7.72 ($\pm$0.37) & 1.44 \\
     &                & R2 & 7.40 ($\pm$0.13) & 7.41 ($\pm$0.14) & 7.45 ($\pm$0.24) & 1.44 \\
AU-2 & 490 (0.92\%)   & R1 & 7.40 ($\pm$0.28) & 7.33 ($\pm$0.27) & 7.21 ($\pm$0.32) & 2.26 \\
     &                & R2 & 7.44 ($\pm$0.25) & 7.32 ($\pm$0.16) & 7.94 ($\pm$1.12) & 1.35 \\
AU-3 & 475 (0.89\%)   & R1 & 8.15 ($\pm$0.26) & 8.04 ($\pm$0.30) & 8.55 ($\pm$1.16) & 2.21 \\
     &                & R2 & 8.09 ($\pm$0.05) & 8.22 ($\pm$0.08) & 16.94 ($\pm$12.78) & 1.32 \\
\hline
\end{tabular}
\end{table}

The correction described in Section~\ref{sec:correction} changes only a small fraction of the training images, which are
blank or mostly missing in V3-2025.R1 and paired with valid labels. Table~\ref{tab:dataset_r2} shows the updated number of TCs and images of the V3-2025.R2 dataset. 
The comparison of the two releases
therefore also serves as an evaluation of how corrupted input images in the training data affect the
benchmark results.

\paragraph{Setup.}
We trained the baseline model of Section~\ref{sec:single-benchmark} (ResNet-18, full 512 $\times$ 512 images,
SS-L2 split, 50 epochs, five runs) on both releases for five single-channel datasets: WP-1, WP-3, AU-1,
AU-2 and AU-3. WP-2, whose corrected images (1,253) are almost the same observations as those of WP-3, was
not included.
The evaluation sets of the two releases are identical. The training sets differ only in the corrected
images and, for AU-1 and AU-2, in the 23 removed GOES-9 images (Table~\ref{tab:r2}).

\paragraph{Evaluation metric.}
Section~\ref{sec:benchmarking} reports the evaluation RMSE at epoch 50. However, the
evaluation RMSE fluctuates from epoch to epoch until the end of training, and for the AU
datasets it occasionally spikes for a single epoch (for example, from 8 to 28--38~hPa) on both releases.
The value at a single epoch therefore depends on where a run happens to end. For this comparison, we use
the median of the evaluation RMSE over the last ten epochs (41--50) of each run, and also report the median
over the last five epochs (46--50) and the value at epoch 50 to show how the conclusions depend on this
choice. Although we used the end of epoch 50 in Section~\ref{sec:benchmarking} to keep the continuity between V1 and V2 benchmarking experiments, we recommend the median over the last ten epochs for future comparisons with the same protocol;
Selecting the epoch on a separate validation set would be a further improvement.

\paragraph{Results.}

The R1 runs in Table \ref{tab:r2} were retrained for this comparison and differ from Tables \ref{tab:north-results-pressure} and \ref{tab:south-results-pressure} (e.g., 6.66 vs 6.56 hPa for WP-1 at epoch 50) because of run-to-run variability and the single-epoch spikes discussed above.

The corrupted images strongly affect the training error: with 0.3--0.9\% corrupted training observations,
the training RMSE at epoch 50 is 2.0--2.9~hPa on V3-2025.R1 and 1.3~hPa on V3-2025.R2, while AU-1, with
only 0.06\%, shows no difference. Because a blank image carries no information about the label, these
observations cannot be fitted and dominate the remaining training error.

The effect on the evaluation RMSE is much smaller. On WP, V3-2025.R2 improves the median over the last ten
epochs by 0.10~hPa for WP-1 and 0.12~hPa for WP-3, larger than the variability across runs. On AU, the
differences ($-$0.13, $+$0.04, and $-$0.06~hPa for AU-1, AU-2, and AU-3) are within the variability across
runs, and their sign depends on the metric. The difference for AU-1, which has only 39 corrupted
observations, illustrates the size of differences that arise from run-to-run variability alone on the
smaller AU evaluation set. The value at epoch 50 is less reliable: the large mean for AU-3 on V3-2025.R2 is
caused by two runs that ended on a spike (28.1 and 33.4~hPa, compared with about 8~hPa in the preceding
epochs).

These results suggest that a small fraction of corrupted input images, paired with valid labels, inflates
the training error but has only a limited effect on the generalization performance of the baseline model,
on the order of 0.1~hPa where it can be detected. The benchmarks of Section~4, computed on V3-2025.R1,
therefore remain representative for V3-2025.R2.

\end{document}